\documentclass[10pt, a4paper]{article}

\usepackage[final]{lrec2026} 
\usepackage{booktabs}
\usepackage{multirow}
\usepackage{array}
\usepackage{amssymb} 
\usepackage{pifont}  
\usepackage{graphicx} 
\usepackage{colortbl}
\usepackage{stfloats}
\usepackage{tabularx}
\usepackage{amsmath}
\usepackage{xcolor}
\usepackage{setspace}
\usepackage{geometry}
\title{M3-SLU: Evaluating Speaker-Attributed Reasoning \\in Multimodal Large Language Models}

\name{Yejin Kwon, Taewoo Kang, Hyunsoo Yoon$^{\dagger}$, and Changouk Kim$^{\dagger}$}

\address{
Department of Industrial Engineering, Yonsei University \\Seoul, Republic of Korea \\
\{beckykwon, hs.yoon, kimco\}@yonsei.ac.kr, gangtaeu02@gmail.com}

\abstract{We present M3-SLU, a new multimodal large language model (MLLM) benchmark for evaluating multi-speaker, multi-turn spoken language understanding. While recent models show strong performance in speech and text comprehension, they still struggle with speaker-attributed reasoning, the ability to understand who said what and when in natural conversations.
M3-SLU is built from four open corpora (CHiME-6, MELD, MultiDialog, and AMI) and comprises over 12,000 validated instances with paired audio, transcripts, and metadata. It includes two tasks: (1) Speaker-Attributed Question Answering and (2) Speaker Attribution via Utterance Matching. We provide baseline results for both cascaded pipelines and end-to-end MLLMs, evaluated using an LLM-as-Judge and accuracy metrics. Results show that while models can capture what was said, they often fail to identify who said it, revealing a key gap in speaker-aware dialogue understanding. M3-SLU offers as a challenging benchmark to advance research in speaker-aware multimodal understanding.
 \\ \newline  }

\begin{document}

\maketitleabstract

\section{Introduction}
Multimodal Large Language Models (MLLMs) have begun to blur the boundary between modalities, that is, seeing, hearing, and reasoning. Advances in models such as Qwen3-Omni \citep{omni2025qwen3}, Gemini 2.5 \citep{gemini_2_5}, and GPT5 \citep{gpt4} have demonstrated how far AI systems can extend their understanding beyond text, seamlessly integrating visual, auditory, and linguistic cues to perceive the world in richer ways. Particularly, Audio-Language Models (ALMs) that integrate auditory representations into large language models, including Qwen2-Audio \citep{qwen2_audio}, Audio Flamingo \citep{flamingo3}, and Voxtral \citep{voxtral}, have recently achieved remarkable progress in bridging speech and language understanding. These models enable machines not only to transcribe speech but also to understand intent, summarize dialogues, and generate coherent responses.

Despite their impressive multimodal capabilities, most existing MLLMs still assume single-speaker conditions, achieving strong performance in Automatic Speech Recognition (ASR) but leaving Speaker Diarization (SD) largely unaddressed \citep{speakerlm}. Yet, real-world conversations are far more complex than such single-speaker settings. Understanding “who spoke when and what” offers a more comprehensive and meaningful perspective on real-world conversations \citep{gao2025multimodal}. In particular, in natural interactions, utterance sequences may be immediately continuous or may overlap temporally; overlapping talk often occurs around turn transitions and is characterized by backchannels, interruptions, and simultaneous first-starts \citep{levinson2015timing, knudsen2020forgotten, schegloff2000overlapping}.

\begin{figure}
\begin{center}
\includegraphics[width=\columnwidth]{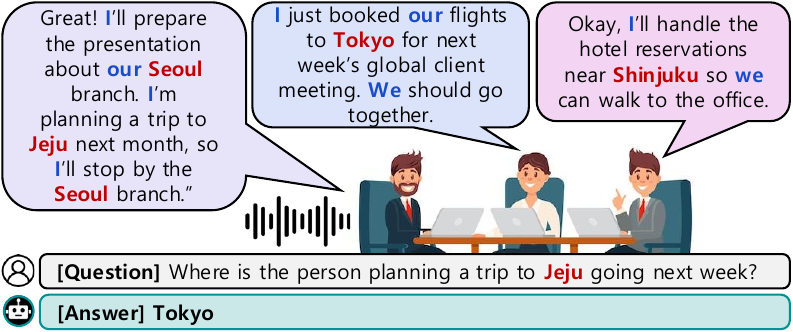}
\caption{Overview of the M3-SLU Benchmark.}
\label{fig.intro}
\end{center}
\end{figure}

\begin{table*}[h]
\centering
\small
\renewcommand{\arraystretch}{1.15}
\resizebox{\textwidth}{!}{%
\begin{tabular}{lccccccc}
\toprule
Benchmarks & SLURP & VoiceBench & MMSU & MMAU & AudioBench  & MSU-Bench & \cellcolor{blue!10}\textbf{M3-SLU (Ours)} \\
\midrule
Speaker-Oriented         & X & X & X & X & X & O & \cellcolor{blue!10}\textbf{O} \\
Multi-speaker            & X & X & X & X & X & O & \cellcolor{blue!10}\textbf{O} \\
Speaker-attributed Reasoning & X & X & X & X & X & $\triangle$ & \cellcolor{blue!10}\textbf{O} \\
Audio Source             & TTS+RPC & TTS+RPC & RPC & RPC & RPC & RPC & \cellcolor{blue!10} \textbf{RPC} \\
Conversation Type        & Monologue & Monologue & Dialogue & Dialogue & Dialogue & Dialogue & \cellcolor{blue!10} \textbf{Dialogue} \\
Conversation Length      & Short & Short & Short & Short & Short & Short & \cellcolor{blue!10}\textbf{Long (Over 1 Min)} \\
\bottomrule
\end{tabular}
}
\caption{Comparison between M3-SLU and existing speech-language understanding benchmarks.}
\label{tab:m3slu_comparison}
\end{table*}

Understanding "Who spoke When and What" in multi-party conversations is a crucial step toward socially intelligent AI. However, most existing speech benchmarks \citep{chen2024voicebench, yang2024air, sakshi2024mmau} address speaker-related and general dialogue tasks together, without separately examining the distinct challenges of speaker-centric understanding in real conversations \citep{wang2025msu}. To close the gap between current MLLM evaluation and real-world conversational complexity, we introduce the \textbf{M3-SLU Benchmark} (\textbf{M}ulti-Speaker, \textbf{M}ulti-Turn, and \textbf{M}ulti-Modal \textbf{S}poken \textbf{L}anguage \textbf{U}nderstanding), as illustrated in Figure~\ref{fig.intro}. Our key contributions:

\begin{itemize}
    \item We constructed M3-SLU benchmark using four open multi-speaker corpora --- \textbf{CHiME-6} \citep{bibil_chime6}, \textbf{MELD} \citep{bibil_meld}, \textbf{MultiDialog} \citep{bibil_multidialog}, and \textbf{AMI} \citep{bibil_ami}, reflecting diverse acoustic conditions and conversational patterns such as overlaps and rapid turns.
    \item We propose the M3-SLU benchmark and \textbf{evaluation framework for MLLMs}, designed to measure performance on \textbf{two simple yet challenging tasks that can only be solved by correctly identifying the speaker}.
\end{itemize}

\section{Related Work}
\label{sec:related_work}

\subsection{Speech Understanding Models}
Speech understanding has advanced beyond mere transcription toward comprehension of spoken language -- capturing both meaning and paralinguistic cues such as prosody and tone. Earlier pipeline systems that linked ASR to NLP models often lost acoustic detail and propagated recognition errors, prompting a shift toward end-to-end architectures that map raw audio directly to semantic representations. This architectural shift has enhanced robustness and enabled deeper speech understanding, as shown by recent models such as SpeechGPT \citep{zhang2023speechgpt}, Salmonn \citep{tang2023salmonn, yu2025salmonn}, Glm-4-voice \citep{zeng2024glm}, and Gemini \citep{team2023gemini}. These models typically follow two main approaches: (1) Using an audio adaptor, as in Audio Flamingo 3 \citep{flamingo3} and Voxtral \citep{voxtral}, or (2) Directly combining an audio encoder with an LLM, as in Qwen2-Audio \citep{qwen2_audio}.

This paradigm has also paved the way for the emergence of more specialized capabilities, such as the Speaker LM \citep{speakerlm}, which captures individual vocal signatures and uses them in reasoning, and the MT-LLM \citep{meng2025large}, designed to disentangle and process dialogue from concurrent speakers. Looking forward, the frontier of research is expanding into multimodal domains where models like GPT-5 \citep{gpt4} and Qwen2.5-Omni \citep{omni2025qwen3} fuse auditory streams with visual data to achieve a contextually richer, more human-aligned interaction.

\subsection{Speech Understanding Benchmarks}

As the performance of Large Audio-Language Models (LALMs) has advanced, developing benchmarks that evaluate their complex speech understanding capabilities has become a major research focus. Early benchmarks in Table\ref{tab:m3slu_comparison} such as SLURP \citep{bastianelli2020slurp} and VoiceBench \citep{chen2024voicebench} primarily focused on single-turn, single-speaker intent classification using synthetic or short speech segments (TTS\footnote{TTS (Text-to-Speech): Speech audio is synthetically generated from written text using a text-to-speech engine.} + RPC\footnote{RPC (Real/Recorded Speech Corpus): Speech audio is naturally recorded from human speakers in real environments.}), laying the groundwork for fundamental speech-language understanding. Subsequent datasets including MMAU \citep{sakshi2024mmau}, MMSU \citep{wang2025mmsu}, and AudioBench \citep{wang2024audiobench} expanded their evaluation scope to multi-turn audio-based question and tasks such as intent classification and emotion interpretation, aiming to assess more nuanced aspects of speech understanding.

However these benchmarks, while often built from long recordings, evaluated only short conversational segments (typically under 30 seconds) and were mainly oriented toward intent, emotion, or speaker recognition rather than context-grounded reasoning. Moreover, they did not address multi-speaker scenarios where understanding requires reasoning over speaker identities and interactions.

\subsection{Multi-speaker Speech Understanding Benchmarks}

As summarized in Table\ref{tab:m3slu_comparison}, MSU-Bench \citep{wang2025msu} marked the first benchmark specifically dedicated to the rigorous evaluation of multi-speaker understanding in realistic conversational scenarios. Yet, it still focused on short dialogues and failed to capture reasoning that spans multiple turns and speakers. Building on this trajectory, we propose the \textbf{M3-SLU} benchmark, designed to assess long-form conversational understanding in segments over one minute long. Unlike prior benchmarks, M3-SLU comprises real multi-speaker dialogues lasting between one and three minutes and features speaker-attributed question answering tasks that require identifying concrete nouns (e.g., objects, places, times, numbers, names) instead of abstract intents or emotions.

\begin{figure*}[t]
    \centering
    \includegraphics[width=\textwidth]{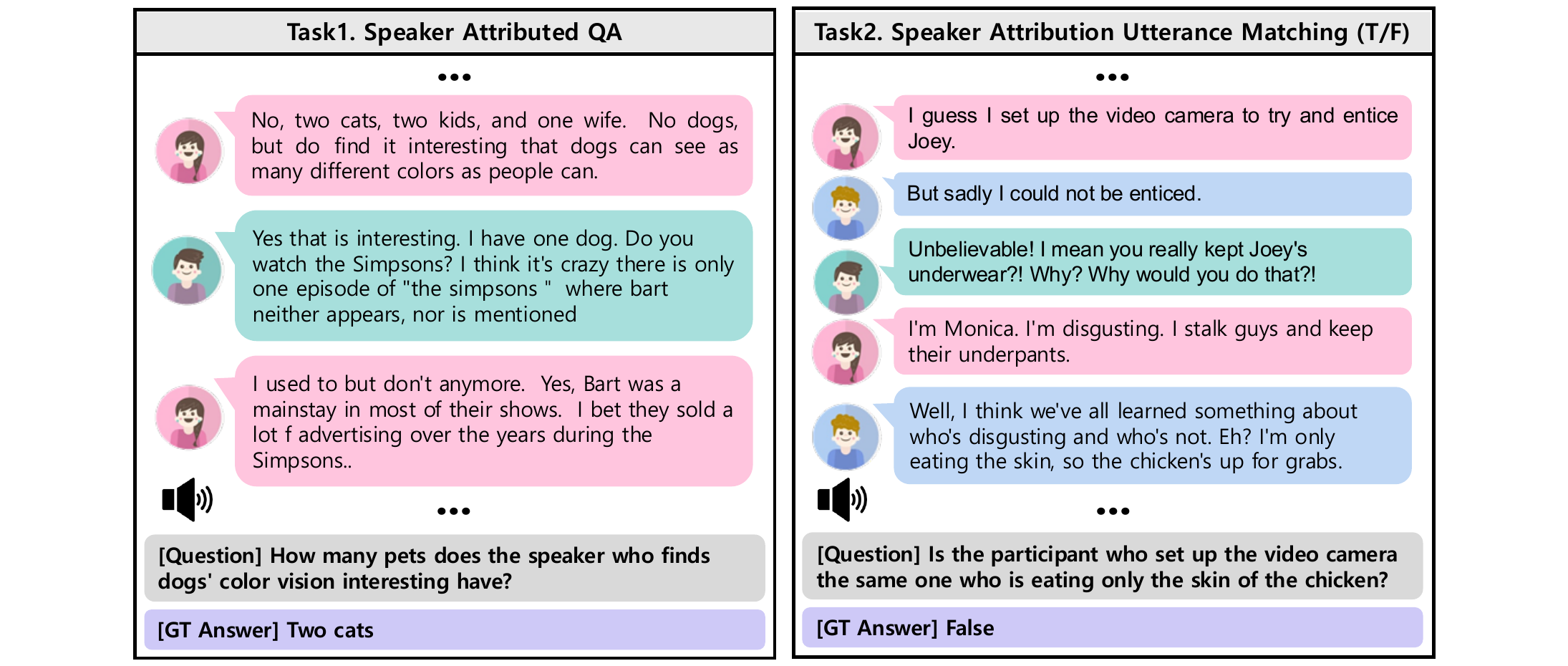}
    \caption{Example of M3-SLU Benchmark. Although they are expressed in text form, they are all voice conversation datasets. The generated Question and GT Answer are in text format.}
    \label{fig:example}
\end{figure*}

\section{M3-SLU Benchmark}

\subsection{Purpose of Benchmarks}

Our M3-SLU benchmark is designed to measure how accurately a model can understand complex multi-speaker conversations and provide appropriate answers to related questions. Therefore, \textbf{we designed tasks that require the MLLM models to listen to the conversation and distinguish between speakers in order to answer correctly}.
As illustrated in  Figure\ref{fig:example}, we propose two tasks: \textbf{Task 1. Speaker-Attributed QA} and \textbf{Task 2. Speaker Attribution Utterance Matching (T/F)}.

\subsection{Overview of Benchmarks}

M3-SLU consists of two core evaluation tasks generated from 4 public multi-speaker dialogue datasets — CHiME-6 \citeplanguageresource{chime6}, MELD \citeplanguageresource{meld}, MultiDialog \citeplanguageresource{multidialog}, and AMI \citeplanguageresource{ami}. \textbf{CHiME-6} consists of real dinner-party recordings captured in noisy environments with overlapping speech and distant microphones, making it ideal for evaluating speech robustness and diarization accuracy. \textbf{MultiDialog} covers multi-topic conversations designed for contextual understanding across diverse domains. \textbf{MELD}, based on the Friends TV series, provides multimodal emotional dialogues that emphasize emotion recognition and sentiment analysis. Finally, \textbf{AMI} includes real business meeting recordings commonly used for summarization and decision-making tasks, capturing realistic multi-speaker interactions in professional settings. (Details and statistics for each dataset are provided in Appendix A.)

\begin{table}[!h]
\centering
\small
\renewcommand{\arraystretch}{1.2}
\resizebox{\columnwidth}{!}{%
\begin{tabular}{lcccc}
\toprule
 & \textbf{CHIME-6} & \textbf{MELD} & \textbf{MultiDialog} & \textbf{AMI} \\
\midrule
Total (h) & 2.91 & 0.44 & 113.04 & 74.87 \\
Seg. Dur (s) & 116.81 & 56.12 & 97.92 & 126.64 \\
Utt./Seg & 66.40 & 13.92 & 15.51 & 87.70 \\
\# of Speakers & 2--4 & 2--8 & 2 & 2--5 \\
\bottomrule
\end{tabular}%
}
\caption{Segment-level statistics of datasets used in M3-SLU, including total audio duration, average segment length, average number of utterances per segment, and the number of speakers.}
\label{tab:dataset_stats}
\end{table}

\begin{table}[!h]
\centering
\small
\renewcommand{\arraystretch}{1.1}
\resizebox{\columnwidth}{!}{ 
\begin{tabular}{lccc}
\toprule
\textbf{Dataset} & \textbf{Task 1 (Q\&A)} & \textbf{Task 2 (T/F)} & \textbf{Total} \\
\midrule
CHiME-6 & 642 & 1,063 & 1,705 \\
MultiDialog & 2,793 & 4,156 & 6,949 \\
MELD & 388 & 597 & 985 \\
AMI & 1,103 & 2,131 & 3,234 \\
\midrule
\textbf{Total} & \textbf{4,926} & \textbf{7,947} & \textbf{12,873} \\
\bottomrule
\end{tabular}
}
\caption{Composition of the M3-SLU Benchmark}
\label{tab:m3slu_final_composition}
\end{table}

\begin{figure*}[!t]
    \centering
    \includegraphics[width=\textwidth]{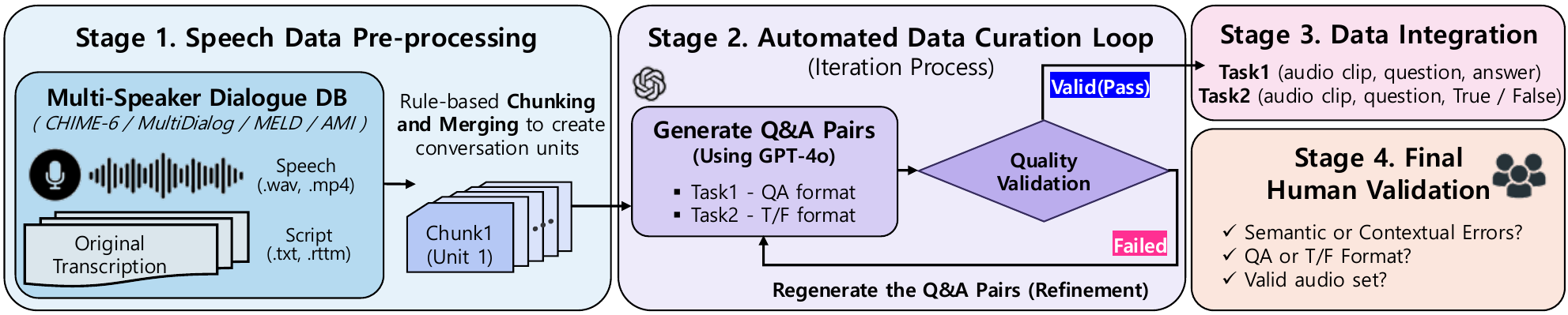}
    \caption{The 4-Stage Hybrid Pipeline for M3-SLU Benchmark Construction.}
    \label{fig:constructi}
\end{figure*}

\begin{figure*}[!t]
    \centering
    \includegraphics[width=\textwidth]{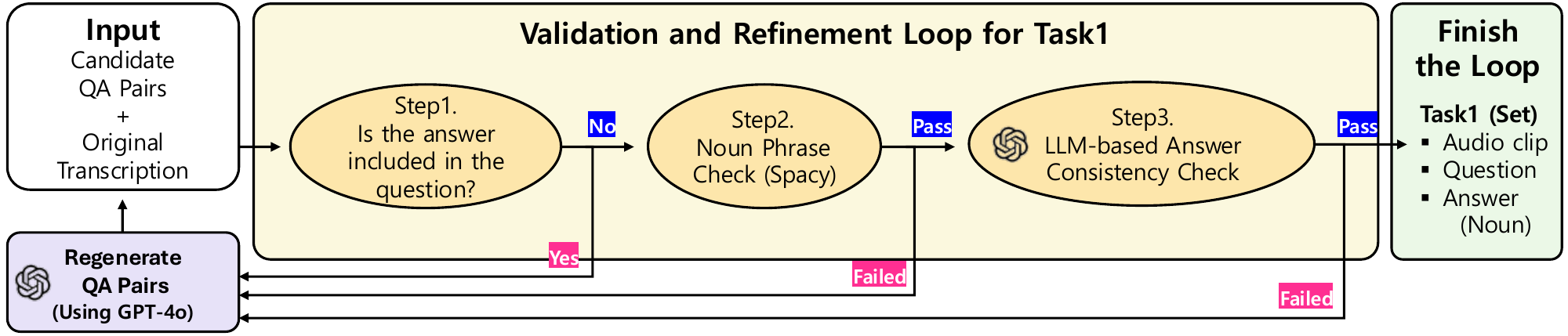}
    \caption{Detailed Pipeline for Validation and Refinement in Task 1(QA)}
    \label{fig:task1}
\end{figure*}

Task 1 (QA) and Task 2 (T/F) were constructed from approximately 8k multi-speaker dialogue segments, each longer than one minute. As shown in Tables \ref{tab:dataset_stats} and ~\ref{tab:m3slu_final_composition}, we finalized 12,873 challenging data instances (segments) for the benchmark, and each instance involves at least two speakers. And, every data instance must consist of at least two speakers. The two proposed tasks are as follows:

\begin{itemize}
    \item \textbf{Task 1. Speaker-Attributed QA}: Evaluates a model’s ability to extract concise noun-phrase answers from conversations, testing how well it links information to the correct speaker.
    \item \textbf{Task 2. Speaker Attribution Utterance Match (T/F)}: Evaluates reasoning about speaker identity by checking if two utterances or actions were made by the same person.
\end{itemize}

\subsection{Benchmark Construction Pipeline} 

The 4-stage hybrid benchmark construction pipeline was implemented like Figure~\ref{fig:constructi}, in which LLM-driven automatic screening and human review were jointly employed to maximize the reliability and accuracy of the generated dataset. 

\subsubsection{Stage 1. Speech Data Pre-processing}
To ensure that the benchmark embodies its core characteristics of multi-speaker and multi-turn conversations, we carefully selected and collected high-quality public datasets(CHIME-6, MultiDialog, MELD, and AMI). Long dialogue recordings, mostly around or over one hour in length, were segmented into semantically coherent conversation units, each lasting between one and three and a half minutes. Only samples containing two or more speakers were retained, resulting in \textbf{refined long multi-speaker conversation chunks} that serve as the raw materials for the second-stage processing engine.

\subsubsection{Stage 2. Automated Data Curation Loop}
This critical engine stage manages data quality systematically by cycling through \textbf{Generation → Validation → Refinement}. Preprocessed conversation units are passed to LLM(GPT-4o) to generate task-specific data, immediately followed by automated verification and refinement cycles.

\paragraph{Generation.}
For both Task 1 and Task 2, around 8,000 multi-speaker conversation chunks are preprocessed. The original transcripts (ground-truth scripts) of these chunks were provided to GPT-4o \citep{hurst2024gpt}, which automatically generated corresponding question and answer pairs: short noun-phrase answers for Task~1 and True/False statements for Task~2. For each task, a small set of manually created QA examples was also supplied as few-shot guidance. 

\begin{figure*}[t]
    \centering
    \includegraphics[width=\textwidth]{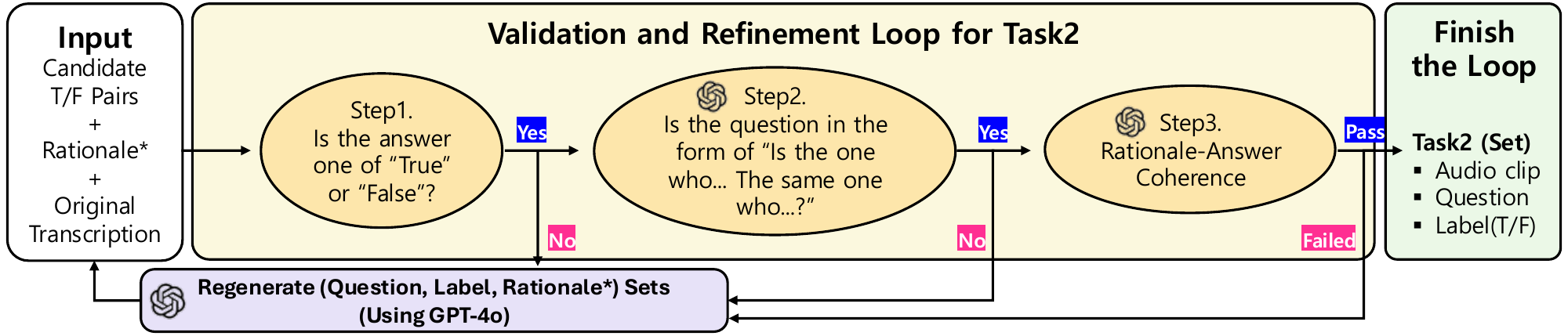}
    \caption{Detailed Pipeline for Validation and Refinement in Task 2(T/F)}
    \label{fig:task2}
\end{figure*}

\paragraph{Validation and Refinement in Task 1.}
After the initial generation stage, an iterative validation and refinement loop in Figure~\ref{fig:task1} was applied to ensure the factual accuracy and linguistic precision of the generated QA pairs. Each iteration involved feeding the original ground-truth scripts and generated QA outputs back into GPT-4o for self-evaluation and regeneration. Through this process, only samples that consistently produced clear, contextually relevant, and noun-phrase-based answers were retained across the entire dataset. 

As shown in Figure~\ref{fig:task1}, \textbf{Step 1} checks whether the answer is redundantly included in the question. \textbf{Step 2} verifies the linguistic validity of the answer as a noun phrase using SpaCy library. \textbf{Step 3} conducts an LLM-based consistency check, in which GPT-4o evaluates whether the generated QA pair is contextually appropriate given the original transcript and whether the provided answer correctly responds to the question. If a candidate pair fails any step, new QA pairs are regenerated by GPT-4o, and the entire validation loop is repeated until all conditions are satisfied.

The iteration stopped after six cycles because the proportion of samples passing Step 2 stayed around 78\%, showing no further improvement. Conversation chunks that repeatedly failed Step 2 were deemed unsuitable for QA generation, as their original scripts lacked extractable noun-phrase answers. These samples were therefore excluded from the dataset. Consequently, the remaining 5,531 candidates advanced to human validation.

\paragraph{Validation and Refinement in Task 2.}
Unlike Task 1, Task 2 additionally required GPT-4o to produce a rationale explaining the reasoning behind each True/False(boolean type) answer. This rationale was used temporarily during the validation process to assess the consistency of each candidate pair, though it was not included in the final benchmark.

The verification process for Task 2 (True/False reasoning) followed the validation and refinement loop illustrated in Figure~\ref{fig:task2}. Each candidate set (question, Boolean label, and rationale) underwent a three-step validation loop. \textbf{Step 1} verified whether the label was a valid Boolean ("True" or "False"). \textbf{Step 2} checked whether the question conformed to the required speaker-attributed form (e.g., “Is the one who … the same one who … ?”), ensuring that the reasoning explicitly involved speaker identity. \textbf{Step 3} evaluated the coherence between the rationale and the answer, filtering out logically inconsistent or semantically irrelevant cases.

Samples that failed any step entered the regeneration step, in which GPT-4o regenerated the candidate set. Each regenerated set was re-validated through the same three steps. After repeating the loop three times, we observed no further increase in the proportion of valid samples. So, we obtained a final set of 8,020 True/False labeled question pairs, each aligned with its corresponding audio clip and ready for subsequent human verification.

\subsubsection{Stage 3. Data Integration}
Text-based data pairs (QA, T/F) that pass automated stages are matched to corresponding original audio clips, packaging both text and speech information into complete multimodal datasets for human annotator use.

\subsubsection{Stage 4. Final Human Validation}
The final and most critical stage ensured the overall integrity and contextual quality of both tasks. In this phase, human annotators manually inspected every remaining instance to identify subtle semantic or contextual errors that automated processes could not capture. For \textbf{Task~1}, annotators carefully examined each question–answer pair to remove cases where the generated noun-phrase answers were vague or linguistically invalid (e.g., phrases like “that thing” or “something”), as well as instances where the question already contained the answer, rendering the question meaningless. For \textbf{Task~2}, reviewers focused on verifying speaker attribution and contextual alignment, discarding samples in which the True/False reasoning relied on incorrect or ambiguous speaker identification.

\section{Evaluation of M3-SLU Benchmark}
Since Task 1 checks noun-phrase matching (QA) and Task 2 judges whether two utterances or actions were made by the same speaker (True/False), we propose distinct evaluation methods for assessing an MLLM’s speaker-attributed reasoning ability.

\paragraph{Task 1 (QA) Evaluation}
Traditional QA evaluation metrics typically rely on Exact Match (EM) and token-level F1 scores. The EM score is the percentage of predictions that match any one of the ground truth answers exactly. The F1 score measures the average overlap between the prediction and ground truth answer \citep{rajpurkar2016squad}. However, these metrics assume textual inputs and do not account for variations arising from speech or pronunciation differences.

\begin{table*}[!t]
\centering
\small
\resizebox{\textwidth}{!}{
\begin{tabular}{lcccccccc}
\toprule
\multirow{2}{*}{\textbf{Model (SD + ASR)}} &
\multicolumn{2}{c}{\textbf{CHiME-6}} &
\multicolumn{2}{c}{\textbf{MELD}} &
\multicolumn{2}{c}{\textbf{MultiDialog}} &
\multicolumn{2}{c}{\textbf{AMI}} \\
\cmidrule(lr){2-3}
\cmidrule(lr){4-5}
\cmidrule(lr){6-7}
\cmidrule(lr){8-9}
 & \textbf{WER} & \textbf{cpWER} & \textbf{WER} & \textbf{cpWER} & \textbf{WER} & \textbf{cpWER} & \textbf{WER} & \textbf{cpWER} \\
\midrule
Pyannote + Whisper-Medium & \underline{0.631} & 0.712 & 0.601 &	0.712 & 0.356 &	0.335 & \textcolor{red}{0.472}	& 0.451 \\
Pyannote + Whisper-Large & 0.635 & 0.713 & 0.604 &	0.707 & 0.391 & 0.354 & \underline{0.487} & 0.478 \\
DiariZen + Whisper-Medium & \underline{0.631} & \textcolor{red}{0.601} & 0.600 & \textcolor{red}{0.581}& \underline{0.355} & \underline{0.162} & \textcolor{red}{0.472} & \textcolor{red}{0.377} \\
(Closed SDR) AssemblyAI & \textcolor{red}{0.532} & \underline{0.631} & \textcolor{red}{0.509} & 0.678 & \textcolor{red}{0.236} & \textcolor{red}{0.157} & 0.531 & 0.472 \\
(Closed SDR) Google STT\textbf{*} & 0.710 &	0.720 & \underline{0.545} & \underline{0.662} & 0.394 & 0.542 & 0.552 & \underline{0.450} \\
\bottomrule
\end{tabular}
}
\caption{Comparison of WER and cpWER across four dialogue audio sets. Both metrics indicate better performance with lower scores. Experiments marked with an asterisk (\textbf{*}) were conducted on 500 randomly sampled instances from each dataset due to research resource limitations.}

\label{tab:wer_cpwer_split}
\end{table*}

So, the new evaluation metric was required for Task 1, which involves listening to speech and identifying the words that appear in the conversation. To address this, we adopted an LLM-as-a-Judge approach, in which GPT-4o evaluated model outputs by considering semantic similarity and phonetic plausibility, a strategy inspired by the evaluation framework in AudioBench \citep{wang2024audiobench}. As shown in Table~\ref{tab:eval_comparison}, this allowed for minor pronunciation- or transcription-related variations—such as “NITE XML” vs. “Night XML”—to be accepted as correct, ensuring a more human-aligned and speech-aware assessment of answer quality. We conducted an LLM-as-a-Judge evaluation using prompts similar to the prompt of \cite{badshah2024reference}. The final evaluation score was defined as the proportion of samples that GPT-4o (LLM-as-Judge) evaluated as Correct among all instances.


\begin{equation}
\text{Score}_{\text{Task1}} = 
\frac{N_{\text{Correct}}^{\text{GPT-4o}}}{N_{\text{Total}}}
\end{equation}

\begin{table}[!h]
\centering
\small
\renewcommand{\arraystretch}{1.2}
\resizebox{\columnwidth}{!}{ 
\begin{tabular}{lccc}
\toprule
\textbf{Case} & \textbf{EM} & \textbf{F1 Score} & \textbf{LLM-as-Judge} \\
\midrule
(GT) \textit{NITE XML} & - & - & - \\
\textit{Nite xml} & Incorrect & Partial & Correct \\
\textit{Night xml} & Incorrect & Partial & Correct \\
\textit{Nite x-m-l} & Incorrect & Incorrect & Correct \\
\bottomrule
\end{tabular}
}
\caption{Comparison of Evaluation Metrics}
\label{tab:eval_comparison}
\end{table}

\paragraph{Task 2 (T/F) Evaluation}
For Task~2, each prediction was evaluated using the standard \textbf{Accuracy} metric, which measures the proportion of correctly classified True/False label.

\section{Experiments and Results}
First, we conducted an Speaker Diarization and Recognition(SDR) Test to verify whether the audio clips in M3-SLU can be accurately transcribed with correct speaker attribution by existing models. Then, we evaluated the M3-SLU benchmark to assess the capability of current E2E MLLMs and cascaded SD + ASR + LLM pipelines in performing speaker-attributed reasoning across multi-speaker dialogues.

\subsection{Experiment Setting for SDR Test}
Following prior work \citep{speakerlm}, we evaluated cascade SD+ASR pipelines on our audio clips in M3-SLU. In particular, we used \textbf{Pyannote 3.1} \citep{bredin2020pyannote} and \textbf{DiariZen} \citep{han2025leveraging} as speaker diarization (SD) modules, both widely recognized for their strong performance on English conversational audio. These were combined with \textbf{Whisper} \citep{radford2023robust} models of varying sizes (Medium and Large) for ASR. To further compare with end-to-end commercial systems, we also included two proprietary SDR models, \textbf{AssemblyAI} and \textbf{Google STT}.

Also, we measured two metrics, average of \textbf{WER} and average of \textbf{cpWER} for audio clips, to assess whether our benchmark achieves proper SDR performance with existing models. WER (Word Error Rate), commonly used in ASR assessment, measures the proportion of word errors between reference and predicted transcripts \citep{morris2004and}. And cpWER (concatenated minimum-permutation WER) adapts WER for multi-speaker data by optimally permuting speakers' transcriptions before scoring \citep{bibil_chime6}.

\subsection{Results of SDR Performance}

Table~\ref{tab:wer_cpwer_split} summarizes the SDR performance of different models across four multi-speaker dialogue audio sets in M3-SLU. Among the open cascade SD + ASR pipelines, \textbf{Diarizen + Whisper-Medium} consistently achieved the most balanced WER and cpWER. Based on this observation, we adopted this Diarizen + Whisper-Medium configuration as our default SD + ASR setting and integrated it with LLMs to conduct the final experiments of our benchmark. Also, closed SDR models such as AssemblyAI achieved further reductions in both WER and cpWER on MultiDialog audio sets. 

Since M3-SLU consists of long segments, rather than the short clips (within 30 seconds) used in previous studies \citep{bibil_chime6, speakerlm}, the reported WER and cpWER are naturally higher. This is because longer segments inherently increase the chance of accumulated transcription errors over time, leading to higher WER, while the frequent speaker transitions in extended dialogues also raise cpWER for each instance.

\subsection{Experiment Setting for M3-SLU Benchmark Evaluation}
To evaluate current models' speaker-attributed reasoning ability on the M3-SLU benchmark, we adopted both cascade (SD + ASR + LLM) and end-to-end (E2E) MLLM methodologies, following prior approaches in spoken language understanding research \citep{speakerlm, wang2025msu}.

In the cascade setting, we first combined speaker diarization (SD) and automatic speech recognition (ASR) models before passing the transcribed text to a large language model (LLM) for question answering. Based on the SDR Test results in Table~\ref{tab:wer_cpwer_split}, the \textbf{Diarizen + Whisper-Medium} combination exhibited consistently balanced multi-speaker transcription performance across our audio datasets, therefore we adopted this combination as the default SD + ASR configuration. The transcribed text was then passed to LLMs such as \textbf{Llama3.1-8B} and \textbf{Mistral-7B/24B} to investigate the impact of LLM scale. We also included commercial \textbf{AssemblyAI}'s transcription result to analyze the difference in transcription text quality.

In parallel, we evaluated end-to-end Speech-LLM and Multimodal LLMs, which directly process raw audio inputs without intermediate transcription. For the Speech-LLM evaluation, we tested \textbf{Qwen2-Audio-7B} and \textbf{Voxtral-Small-24B}, while the Multimodal LLM evaluation included \textbf{Qwen2.5-Omni-7B} and \textbf{Qwen3-Omni-30B}, which jointly handle audio and textual reasoning in a unified framework.

\definecolor{headergray}{gray}{0.9}
\definecolor{sectionpurple}{RGB}{245,240,255} 

\begin{table}[t]
\centering
\small
\renewcommand{\arraystretch}{1.2}
\resizebox{\columnwidth}{!}{ 
\begin{tabular}{lcc}
\toprule
\textbf{M3-SLU Benchmark} & \textbf{Task1} & \textbf{Task2} \\
\midrule
\rowcolor{sectionpurple}
\multicolumn{3}{l}{\textbf{SD + ASR + LLM}} \\
\specialrule{0.4pt}{0pt}{0pt}
\textit{GT Script(Gold)+Llama3.1(8B)} & 0.9577 & 0.5787 \\
\textit{GT Script(Gold)+Mistral(7B)} & 0.8717 & 0.5409 \\
Diarizen+whisper+Llama3.1(8B) & 0.7863 & 0.5620 \\
AssemblyAI+Llama3.1(8B) & 0.9192 & 0.5452 \\
Diarizen+whisper+Mistral(7B) & 0.7665 & 0.5490 \\
Diarizen+whisper+Mistral(24B) & 0.8068 & 0.6544 \\
\specialrule{0.5pt}{0pt}{0pt}

\rowcolor{sectionpurple}
\multicolumn{3}{l}{\textbf{E2E Speech LLM}} \\
\specialrule{0.4pt}{0pt}{0pt}
Qwen2-Audio(7B) & 0.0602 & 0.4960 \\
MistralAI-Voxtral(24B) & 0.8375 & 0.5169 \\
\specialrule{0.5pt}{0pt}{0pt}

\rowcolor{sectionpurple}
\multicolumn{3}{l}{\textbf{E2E Multimodal LLM}} \\
\specialrule{0.4pt}{0pt}{0pt}
Qwen2.5-Omni(7B)  & 0.6883 & 0.5071 \\
Qwen3-Omni(30B) & 0.7762 & 0.5760 \\
\bottomrule
\end{tabular}}
\caption{M3-SLU benchmark results comparing cascaded (SD+ASR+LLM) and E2E Speech/Multimodal LLM models. For Task 1, the evaluation scores were obtained using the LLM-as-Judge approach, while for Task 2, the values represent accuracy based on the correctness of True/False judgments. Both scores are higher-the-better metrics.}
\label{tab:final_result}
\end{table}

\subsection{Results of M3-SLU Benchmark}
The top two scores in Table~\ref{tab:final_result} correspond to the \textit{GT Script(Gold)} configuration, where the model was provided with the original ground-truth transcription that had been perfectly segmented by speakers.

\paragraph{Evaluation on Task 1} 
As expected, when the \textit{GT Script(Gold)} was provided to the LLM, this configuration achieved the highest score, since the model received perfectly transcribed and speaker-segmented text, effectively removing any noise or cascading errors during the SD and ASR stages. In addition, the cascade setting (SD + ASR + LLM) also demonstrated reasonably strong QA performance, indicating that despite inevitable errors from the SD + ASR modules, the overall pipeline was still able to preserve a substantial amount of semantic and speaker-related information.

In the upper part of Table~\ref{tab:final_result}, the results demonstrate how the quality of the SD + ASR pipeline directly influences the final QA performance in the cascade setting. Although both settings use the same LLM (Llama 3.1-8B), AssemblyAI + Llama 3.1-8B achieves 0.9192, substantially outperforming Diarizen + Whisper + Llama 3.1-8B (0.7863). As shown in Table~\ref{tab:wer_cpwer_split}, AssemblyAI exhibits remarkably higher transcription accuracy, particularly on the MultiDialog Audio dataset, showing a clear margin over the Diarizen + Whisper-Medium results. That is, \textbf{more precise transcriptions and speaker labels allow the LLM to better understand who said what}, thereby yielding QA results nearly comparable to those obtained with gold transcriptions (GT Script + LLM). Also, the performance difference between Mistral-7B (0.7665) and Mistral-24B (0.8068) shows that increasing the model size leads to a moderate improvement in Task 1 results, suggesting that \textbf{larger LLM better leverage the transcribed and diarized input for understanding speaker-attributed content}.

In the lower part of Table~\ref{tab:final_result}, E2E Speech-LLM and Multimodal LLM models show relatively lower performance on Task 1, compared to the cascade setting. Only the larger models, such as MistralAI-Voxtral (24B) and Qwen3-Omni (30B), achieved scores approaching 80\%, indicating that \textbf{our M3-SLU Task 1 remains a challenging problem for current E2E Speech-LLMs and MLLMs}.

\paragraph{Evaluation on Task 2} As shown on the right side of Table~\ref{tab:final_result}, the Task 2 results reveal that no model configurations exceeded 70\%, a surprisingly low performance considering that the Task 2 is binary (True/False) classification. This suggests that \textbf{accurate speaker-attributed utterance matching is impossible for both cascade and E2E models under the current framework}. Even in the cascade setting with gold transcripts provided, the models failed to accurately distinguish speakers. The best result was obtained with the Diarizen + Whisper + Mistral (24B) combination, which achieved only 0.6544 on Task 2.

As shown in the comparison between Task 1 and Task 2 results in Table \ref{tab:final_result}, models demonstrated a \textbf{noticeable gap between understanding what was said and who said it}. While Task 1 could be partially solved by leveraging contextual cues without explicit speaker distinction, Task 2 inherently required precise speaker identification to match utterances correctly. This means that \textbf{although current models can comprehend the content of conversations, they remain largely incapable of reasoning about speaker attribution}, highlighting the persistent gap toward true multi-speaker understanding. Therefore, advancing and evaluating future MLLMs will require our M3-SLU benchmark as a foundation for genuine multi-speaker understanding.

\paragraph{Evaluation in Closed Models}

As shown in Table~\ref{tab:closed} below, current commercial models such as GPT-4o-Audio and Gemini-2.5-Flash-Audio completely fail to perform multi-speaker understanding, indicating that they are still unable to distinguish and reason over different speakers in conversational audio.

\begin{table}[!h]
\centering
\small
\renewcommand{\arraystretch}{1.2}
\resizebox{\columnwidth}{!}{ 
\begin{tabular}{lcc}
\toprule
\textbf{M3-SLU Benchmark} & \textbf{Task1} & \textbf{Task2} \\
\midrule
\rowcolor{sectionpurple}
\multicolumn{3}{l}{\textbf{Closed Models}} \\
\specialrule{0.4pt}{0pt}{0pt}
GPT-4o-Audio* & 0.32 & 0.51 \\
Gemini-2.5-Flash-Audio* & 0.41 & 0.54 \\
\bottomrule
\end{tabular}}
\caption{M3-SLU benchmark results using closed commercial models. Due to limited research resources, both models were evaluated on a randomly selected 100 samples.}
\label{tab:closed}
\end{table}

\subsection{Human Verification of LLM-as-Judge Evaluation}
Since M3-SLU Task 1 involves predicting noun phrases from audio inputs, we adopted an LLM-as-a-Judge evaluation method using GPT-4o. To verify its reliability, we manually compared GPT-4o’s judgments with human judgments on 200 randomly selected samples. Specifically, we used GT noun answers and predictions from the Diarizen + Whisper-Medium + LLaMA 3.1 (8B) experiment. The results showed 96.5\% agreement between GPT-4o and human evaluators, demonstrating that our LLM-as-Judge evaluation is consistent with human judgment and not arbitrarily biased.

\section{Conclusion}
We introduced M3-SLU, a benchmark that reveals a key limitation of current MLLMs—the inability to comprehend "who spoke when and what" in long multi-speaker dialogues. Through two targeted tasks, Speaker-Attributed QA and Utterance Matching, M3-SLU isolates the challenge of speaker reasoning beyond simple transcription. Our experiments show that while existing cascaded pipelines and MLLMs can capture what was said, they consistently fail to track who said it, even with accurate transcripts. This underscores a critical gap in speaker attribution and multi-speaker reasoning. Building on real, naturally occurring conversations with speaker-attributed annotations, M3-SLU offers a structured evaluation setting that addresses the limitations of synthetic or short-turn benchmarks. The consistently low performance of current state-of-the-art models across both tasks reflects the complexity of speaker-grounded reasoning, which is unlikely to be resolved through scaling alone. M3-SLU offers a practical testbed for studying how multimodal language models handle multi-speaker conversations in realistic settings. We anticipate that it will guide the development of modeling strategies, evaluation methods, and training practices that explicitly incorporate speaker roles, turn-taking, and conversational structure, all of which are essential to dialogue comprehension.

\section{Limitation}
Our current benchmark is primarily focused on English conversational data. Future work could expand M3-SLU to include a wider range of languages and even more complex, overlapping speech scenarios to further probe the robustness of MLLMs.
In addition, our evaluation framework currently relies on GPT-4o as an LLM-as-Judge to assess model outputs. While this approach enables flexible and semantic-level evaluation, it may overlook subtle variations that arise from speech input, such as minor pronunciation or transcription differences. We are actively exploring more refined evaluation strategies that can account for these speech-induced variations while maintaining fairness and consistency across models.

\section{Ethical Consideration}
All audio data used in this study are sourced from publicly available corpora (CHiME-6, MELD, MultiDialog, and AMI) that provide appropriate research licenses. No private or personally identifiable information (PII) was included. The dataset construction and experiments fully comply with the ethical use policies of the original sources.


\section{Bibliographical References}\label{sec:reference}

\bibliographystyle{lrec2026-natbib}
\bibliography{lrec2026-example}

\section{Language Resource References}
\label{lr:ref}
\bibliographystylelanguageresource{lrec2026-natbib}
\bibliographylanguageresource{languageresource}

\end{document}